\documentclass{article}

\usepackage{arxiv}

\usepackage[utf8]{inputenc} %
\usepackage[T1]{fontenc}    %
\usepackage{hyperref}       %
\usepackage{url}            %
\usepackage{booktabs}       %
\usepackage{amsfonts}       %
\usepackage{nicefrac}       %
\usepackage{microtype}      %
\usepackage{lipsum}
\usepackage{graphicx}
\usepackage{caption}
\usepackage{biblatex}
\usepackage{array}
\usepackage{gensymb}
\usepackage[table]{xcolor}

\title{Deep Image Orientation Angle Detection}

\author{
  Subhadip Maji\thanks{GitHub Repo: \emph{https://github.com/pidahbus}} \\
  M.Tech QROR-II \\
  Indian Statistical Institute, Kolkata \\
  Kolkata, 700108 \\
  \texttt{qr1705@isical.ac.in} \\
   \And
 Smarajit Bose \\
  Interdisciplinary Statistical Research Unit \\
  Indian Statistical Institute, Kolkata \\
  Kolkata, 700108 \\
  \texttt{smarajit@isical.ac.in} \\
}
\DeclareUnicodeCharacter{2212}{-}
\begin{document}
\maketitle

\begin{abstract}
Estimating and rectifying the orientation angle of any image is a pretty challenging task. Initial work used the hand engineering features for this purpose, where after the invention of deep learning using convolution based neural network showed significant improvement in this problem. However this paper shows that with the combination of CNN and a custom loss function specially designed for angles lead to state-of-the-art result. This includes the estimation of orientation angle of any image or document at any degree (0 to 360\degree)

\end{abstract}

\keywords{Image Orientation Angle Detection \and Convolutional Neural Network \and Deep Learning \and Angle Loss Function}

\section{Introduction}

Image orientation angle detection is a pretty challenging task for a machine, because the machine has to learn the features of an image in such a way so that it can detect the arbitrary angle, the image is rotated. Though there are some modern cameras with inertial sensors can correct image orientation in 90 degrees step, but this function generally is not used. In this paper, we proposed a method to detect the orientation angle of a captured image: a post processing step captured in any camera (both older and newer camera models) with any tilted angle (between 0 and 359 degree). After we detect the orientation angle we reverse the angle to correct the orientation of the image. 

From human perspective it is somehow easy to approximately tell the orientation angle of an image based on the elements present in the image. But for a machine an image is just a matrix with pixel values. Thanks to Convolutional Neural Networks for which it has been possible to build an Image Orientation Angle Detection Model which predicts the orientation angle so accurately that it outperforms all the image orientation techniques published in the community.

Orientation correction work has been done since long time for document analysis\cite{cite1, cite2, cite3, cite4, cite6}. These methods need the special structure of the documents images i.e. precise shape of letters or text layout in lines. But for natural images there is no such boundaries available, so it is quite hard for these methods to work properly on the natural images. 

However, this paper is inspired from the work of Fischer et al.\cite{cite7} on image orientation angle estimation for natural images. So, for the most of the parts of this paper, we did a comparative study with Fischer et al.\cite{cite7} and finally with latest CNN architecture, modified loss function and optimizing technique on the same COCO dataset\cite{cite8} we resulted a better model which outperforms Fischer et al.\cite{cite7} with quite a significant margin and our method gives the state-of-the-art result on this problem.

Wei et al.\cite{cite9} used interpolation artifacts by applying rotation to the digital images. However, this method does not work for those images which were not taken upright. Solanki et al.\cite{cite10} predicted the rotation of the printed images by analyzing the pattern of printer dots. But, this method does not work on the digital images.  

Horizon detection\cite{cite11, cite12} is a special kind of image angle detection method but it strongly depends on the presented horizon of the image. However, most of the images do not contain horizon. 

Some works on image orientation angle detection have been termed as a classification tasks\cite{cite13, cite14} where the estimated angle is categorical variable with four to six restricted categories. However, our problem is quite different and can be termed as a regression problem where from 0 to 359 degrees, any angle can be predicted from the given input image with the help of our Orientation Angle Detection Model. 

There have been some works done on the orientation angle detection of only the face images\cite{cite15, cite16, cite4}. But our work has been generalized to any natural images.

\section{Experimental Setup}
\subsection{Data}
\begin{figure}
    \centering
    \includegraphics[scale=0.85]{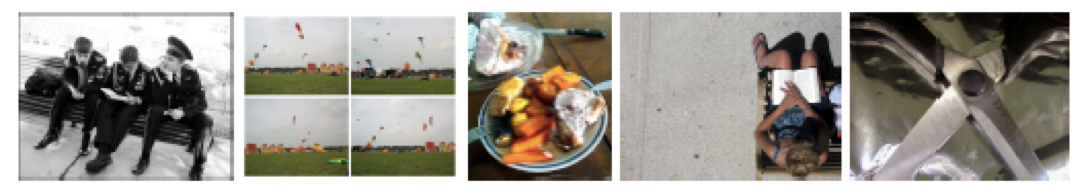}
    \caption{Sample of the discarded images from both validation and test dataset. Reasons for elimination are: slanted images, framed images, undefined oriented images, etc.}
    \label{fig:sample_imgs}
\end{figure}

Ideally for this experiment data can be collected using a camera with a sensitive tilt sensor like an accelerometer. But this process makes data collection much time consuming and expensive. For this reason, we selected Microsoft COCO dataset\cite{cite8} assuming that all the images are orientated correctly i.e. 0-degree rotation. Then we applied artificial rotation to the images and split them into train, validation and test dataset. We selected 2293 images in the validation set, 1000 images in the test set after discarding images which has no orientation because the image was taken from the top, already tilted initially etc. Sample of the discarded images are shown in the Figure \ref{fig:sample_imgs}. 

\subsection{Tasks and Networks}
\begin{figure}[ht]
    \centering
    \includegraphics{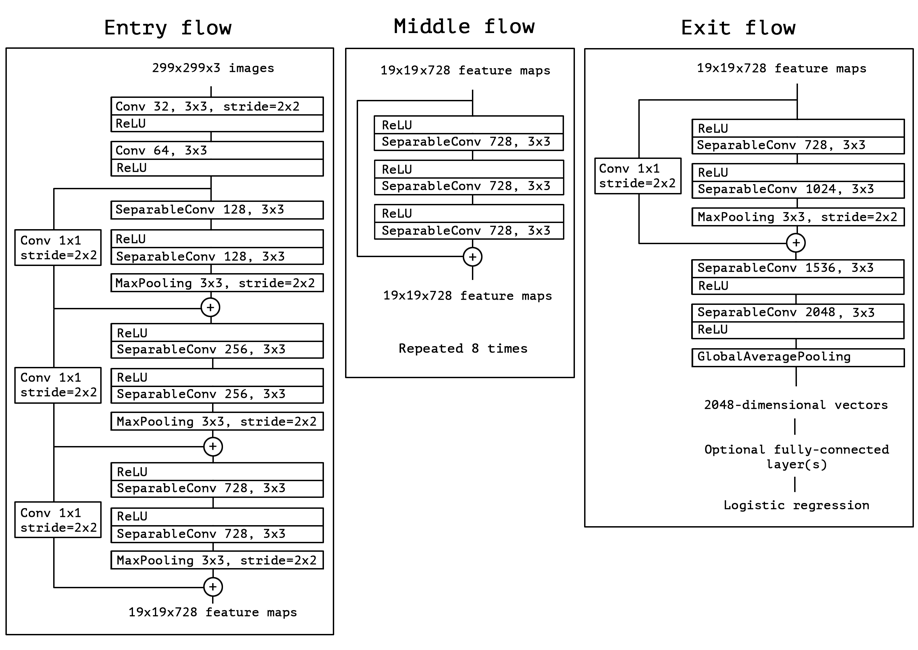}
    \caption{The Xception architecture\cite{cite18}}
    \label{fig:xception}
\end{figure}

Like Fischer et al.\cite{cite7} we too developed our orientation estimation system in three difficulty levels: (1) all the images are rotated randomly within $\pm30$,  (2) all the images are rotated randomly within $\pm45$ and (3) all the images are fully rotated randomly i.e. between 0 to 359 degrees. We name our models for training and estimating images of these three levels as OAD-30, OAD-45 and OAD-360. For the above tasks to perform we have tried all the standard network architectures and among them Xception neual network architecture from Chollet\cite{cite18} turned out to be the best. Figure \ref{fig:xception} shows the Xception architecture in a flowchart. 

\begin{figure}[ht]
    \centering
    \includegraphics{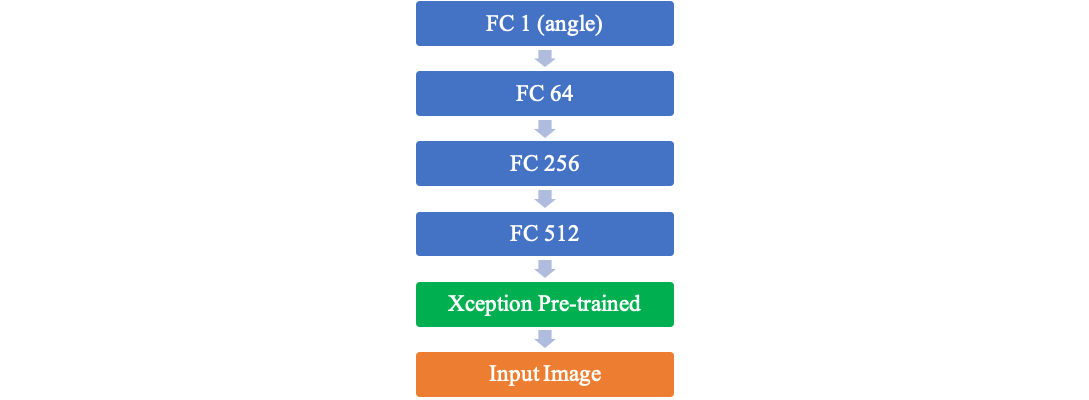}
    \caption{Architecture used for transfer learning model\cite{cite19}}
    \label{fig:architecture}
\end{figure}

Using ImageNet pre-training worked better than training the network from scratch despite the good availability of training data for our tasks. It seems that the class labels from ImageNet help learn semantic features that are useful for the task but too difficult for the network to learn from the orientation objective. Among all the models Xception outperforms others. We extracted the convolution base from the original Xception architecture and added 3 Fully Connected layers (FC) of size 512, 256 and 64 respectively with ReLU activation units. Finally, we added the prediction layer of node size 1 with linear activation unit to predict the image orientation in degree formulating it as a Regression problem as shown in Figure \ref{fig:architecture}. 

\subsubsection{Custom Loss Function:}
Though to detect angle of images randomly oriented within $\pm30$ and $\pm45$ traditional L1 loss can be used, but to detect angle of images randomly oriented between 0 to 359 degrees, the traditional L1 loss will be somewhat misleading. For example, let's assume the true orientation angle of an image is 1\degree and the model predicts the angle to be 359\degree. Then ideally the absolute loss should be 2\degree, but the L1 loss will give the loss = |359-1| = 358\degree. This means ideally model gave good result, but the traditional L1 loss will confuse the model and thus will hinder it's learning. This motivates us to use the below loss function:

\texttt{If the true orientation angle of the image = $t (0 <= t <= 360)$, \\
\quad \quad and predicted orientation angle by the model = $p (0 <= p <= 360)$, \\
then Loss for Image i, $L_i = min\{|t-p|,360-|t-p|\}$ \\
\newline
Overall Loss, $L = mean(L_i)$} \newline

Different types of Optimization Methods have been tried on different types of ImageNet pre-trained models. Some architectures were also initialized from the scratch. It had been clearly seen that pre-trained weights converge better and faster. Also, among the optimization methods Adadelta\cite{cite20} clearly outperforms others. With our custom loss function using Adadelta as optimization method with learning rate 0.1 and Xception pre-trained weights we got the validation absolute angle error to be 1.52, 1.95 and 8.38 for the three difficulty levels mentioned above. These results turned out to be significantly better with respect to Fischer et al.\cite{cite7}. The rest parameters of the Adadelta optimizers were default values set in Keras. 

\subsection{Baseline Methods}
Fischer et al.\cite{cite7} in 2015 did not find any prior work regarding orientation angle estimation of natural images. Hence, for comparison purpose they chose three baseline methods built upon the “Straighten image” function from Matlab Central\cite{cite21} and used two computer vision techniques: Hough transform and Fourier transform. From 2015 to today we too did not find any work in this area. So, to test the performance of our model, we are comparing all the methods described in Fischer et al.\cite{cite7} (section 3.3) with ours.

\section{Results}
\begin{figure}
    \centering
    \includegraphics[scale=0.85]{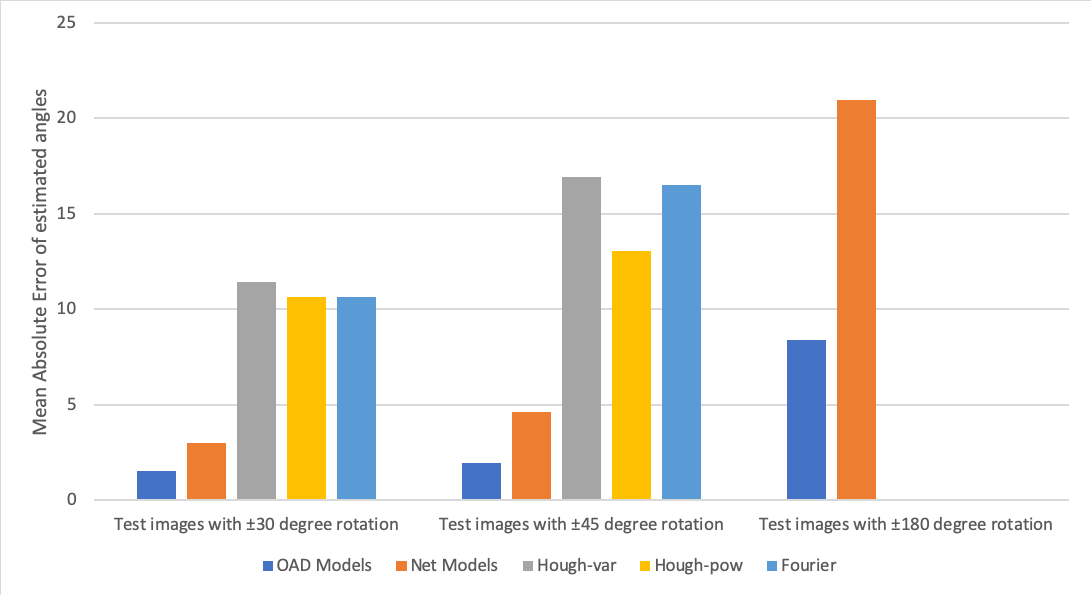}
    \caption{The comparison between our OAD model and other image orientation angle estimation techniques. For estimating orientation angle of test images with $\pm30$-degree rotation we used only OAD-30 model and picked the result of Net-30 from Fischer et al. [7].  Similarly, for estimating rotation of test images with $\pm45$-degree we have used OAD-45 and Net-45 models. Hough and Fourier transformation cannot estimate orientation angle for test images $\pm180$-degree precisely. So, we omitted their results for test images with $\pm180$-degree rotation.}
    \label{fig:result_fig}
\end{figure}

Figure \ref{fig:result_fig} and Table \ref{tab:compare} shows this comparison between our OAD model and other image orientation techniques.

\begin{table}[ht]
    \centering
    \begin{tabular}{|p{2.2cm}|c|c|c|c|c|c|p{1cm}|p{1cm}|c|}
        \hline
        \rowcolor{lightgray} \textbf{Task} & \textbf{OAD-30} & \textbf{Net-30} & \textbf{OAD-45} & \textbf{Net-45} & \textbf{OAD-360} & \textbf{Net-360} & \textbf{Hough-var} & \textbf{Hough-pow} & \textbf{Fourier} \\
        \hline
        Test images with $\pm30$ degree rotation & \textbf{1.52} & 3 & - & - & - & - & 11.41 & 10.62 & 10.66 \\
        \hline
        Test images with $\pm45$ degree rotation & - & - & \textbf{1.95} & 4.63 & - & - & 16.92 & 13.06 & 16.51 \\
        \hline
        Test images with $\pm180$ degree rotation & - & - & - & - & \textbf{8.38} & 20.97 & - & - & - \\
        \hline
    \end{tabular}
    \caption{The comparison between our OAD model and other image orientation angle estimation techniques. It is clearly seen that our OAD model outperforms other baseline methods and achieve very good results.}
    \label{tab:compare}
\end{table}

\begin{figure}
    \centering
    \includegraphics[height=10cm, width=14cm]{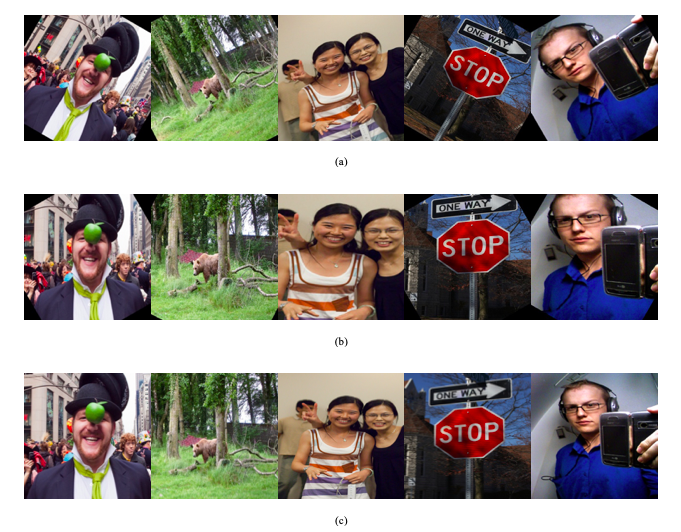}
    \caption{(a) Input Image (b) OAD-30 Model prediction (c) Ground Truth. These are some examples of results from the OAD-30 Model output. Some images are cropped to remove the black portion due to rotation.}
    \label{fig:oad30}
\end{figure}

\begin{figure}
    \centering
    \includegraphics[height=10cm, width=14cm]{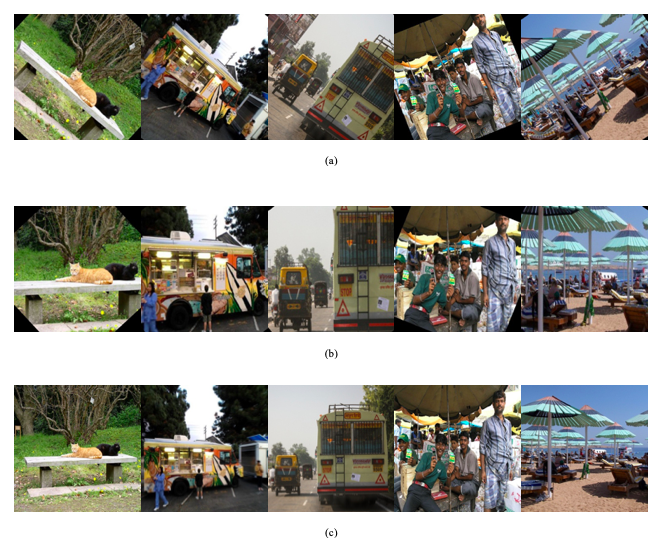}
    \caption{(a) Input Image (b) OAD-45 Model prediction (c) Ground Truth. These are some examples of results from the OAD-45 Model output. Some images are cropped to remove the black portion due to rotation.}
    \label{fig:oad45}
\end{figure}

\begin{figure}
    \centering
    \includegraphics[height=10cm, width=14cm]{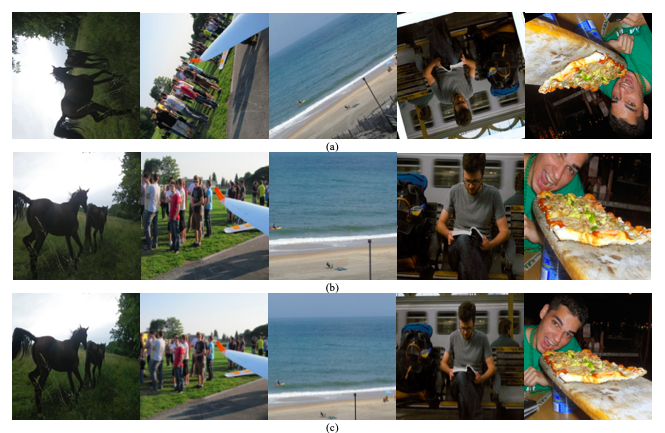}
    \caption{(a) Input Image (b) OAD-360 Model prediction (c) Ground Truth. These are some examples of good results from the OAD-360 Model output. Some images are cropped to remove the black portion due to rotation.}
    \label{fig:oad360good}
\end{figure}

\begin{figure}
    \centering
    \includegraphics[height=10cm, width=14cm]{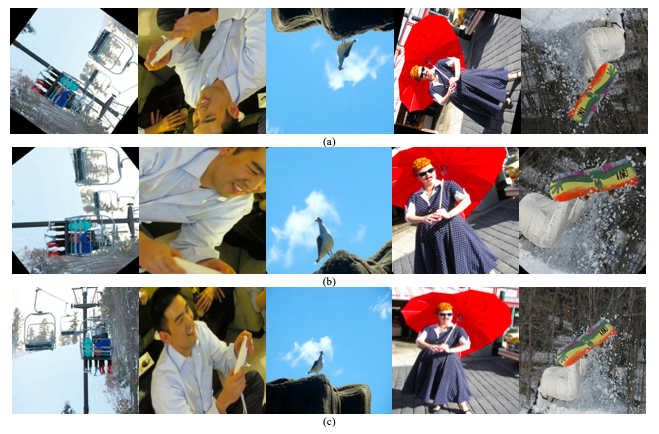}
    \caption{(a) Input Image (b) OAD-360 Model prediction (c) Ground Truth. These are some examples of failed results from the OAD-360 Model output. It is seen that some of the above results are hard for humans also to decide the correct orientation angle. Some images are cropped to remove the black portion due to rotation}
    \label{fig:oad360bad}
\end{figure}

We evaluated our models in test set for three different levels shown in the above graph and table. Here, in this section Figure \ref{fig:oad30}, Figure \ref{fig:oad45} and Figure \ref{fig:oad360good} and \ref{fig:oad360bad} show orientation angle adjustment by our OAD models for three levels respectively. Orientation angle adjustment means predicting the orientation angle of a rotated image and reversing that angle to correct the orientation of the image. For Test images with $\pm30$ and $\pm45$-degree rotation, we find the MAE to be very low (around 2-degree). So orientation of all the test images of these two levels are predicted roughly correctly. But, predicting the orientation angle of randomly fully rotated images is a tough job and as a result we got comparatively worse MAE (8.38-degree). For this reason, figure 7 presents two sets of OAD-360 prediction results: one where the model performed very good and another where it performed comparatively bad. 

\section{Further Analysis on the Improvement}
This section analyses the improvement over Fisher et al.\cite{cite7}. Two major changes with respect to their paper are: 

\begin{itemize}
    \item We used Xception as our pre-trained network where Fisher et al.\cite{cite7} used AlexNet\cite{alexnet}.
    \item We used one custom loss function where they used the traditional L1 loss.
\end{itemize}

So, one argument arises that which factor has more impact on this improvement. To counter this, we did the same experiment with two combinations. One with the Xception Network and traditional L1 loss and another is with the AlexNet Network and with our custom loss. Table \ref{tab:further_analysis} populates the result.

For the task 1 and 2, as the model trains on angles of images between $\pm30$ or $\pm45$, the chance of the model to predict one angle above 180 or below -180 is very rare. In that case, our custom loss function becomes identical to the traditional L1 loss function. So, the Xception with L1 loss model will become OAD-30/OAD-45 and AlexNet with custom loss model will become Net-30/Net-45. This can be seen from the results in the table.

As discussed in section 2.2.1, the scenario is completely different when images are orientated from 0\degree to 359\degree. From the table it is seen that with improved pre-trained model Xception and traditional L1 loss, we get a very little improvement over the work of Fischer et al.\cite{cite7}. But with AlexNet which Fisher et al.\cite{cite7} and our custom loss function we get a very good improvement (though not the best). This proves that this custom loss function has a very powerful impact on the improvement and this loss function can be used for any angle related work in future.

\begin{table}[ht]
    \centering
    \begin{tabular}{|p{3cm}|p{2cm}|p{2cm}|c|c|}
        \hline
        \rowcolor{lightgray} \textbf{Task} & \textbf{OAD-30} & \textbf{Net-30} & \textbf{Xception with L1} & \textbf{AlexNet with Custom loss}\\
        \hline
        Test images with $\pm30$ degree rotation & 1.52 & 3 & 1.55 & 3.09 \\
        \hline
        
    \end{tabular}

    \noindent\begin{tabular}{|p{3cm}|p{2cm}|p{2cm}|c|c|}
        \hline
        \rowcolor{lightgray} \textbf{Task} & \textbf{OAD-45} & \textbf{Net-45} & \textbf{Xception with L1} & \textbf{AlexNet with Custom loss}\\
        \hline
        Test images with $\pm45$ degree rotation & 1.95 & 4.63 & 1.96 & 4.61 \\
        \hline
    \end{tabular}
    
    \noindent\begin{tabular}{|p{3cm}|p{2cm}|p{2cm}|c|c|}
        \hline
        \rowcolor{lightgray} \textbf{Task} & \textbf{OAD-360} & \textbf{Net-360} & \textbf{Xception with L1} & \textbf{AlexNet with Custom loss}\\
        \hline
        Test images with $\pm180$ degree rotation & 8.38 & 20.97 & 18.57 & 10.08 \\
        \hline
    \end{tabular}
    \caption{Further analysis on the improvement to check which factor is more important: improved pre-trained networks or custom loss function?}
    \label{tab:further_analysis}
\end{table}

\section{Conclusion}
This paper shows that with a pre-trained model and a custom loss function state-of-the-art result on image orientation angle detection can be achieved. The custom loss function is a very important part of this paper. Using this loss function any angle related work can be done in future.

\clearpage
\printbibliography
\end{document}